\documentclass[11pt,a4paper]{article}

\usepackage[hyperref]{emnlp-ijcnlp-2019}
\usepackage{times}
\usepackage{latexsym}
\usepackage{xcolor}
\usepackage{url}
\usepackage[utf8]{inputenc}

\aclfinalcopy 

\def\pharmaconer/{\textsf{\small\textsc{PharmaCoNER}}}
\newcommand{\plus}{{\footnotesize\texttt{+}}}
\def\snomed/{\textsc{snomed ct}}
\def\arborex/{{\small\textsf{ARBOREx}}}
\def\AG/{\textit{AnnotGuide}}
\def\HA/{\textsf{\textsc{ha}}}
\def\HAD/{\textsf{\textsc{had}}}
\def\GD/{\textsf{\textsc{gd}}}
\def\sctid/{{\small\textsf{sctid}}}
\def\NER/{\textsf{\textsc{ner}}}
\def\NE/{\textsf{\textsc{ne}}}

\begin{document}
\title{Annotating and normalizing biomedical {\textsf{\textsc{ne}{\normalsize{s}}}} with limited knowledge\thanks{This paper should have been published in the \textit{Proceedings of the 5th Workshop on BioNLP Shared Tasks}. Unfortunately, due to their complete lack of funding, the authors could not afford the registration fees, a mandatory expense for a contribution to be published in the aforementioned proceedings.}}
%
%
\author{Fernando Sánchez León \\
\textsl{unaffiliated} \\
{\small \tt f.sanchez.lcmcvp@gmail.com} \\\And
Ana González Ledesma \\
\textsl{unaffiliated} \\
{\small \tt ana.gonzalez.ledesma@protonmail.com}}

\date{}

\maketitle              
\begin{abstract}
Named entity recognition (\NER/) is the very first step in the linguistic processing of any new domain. It is currently a common process in BioNLP on English clinical text. However, it is still in its infancy in other major languages, as it is the case for Spanish. Presented under the umbrella of the \pharmaconer/ shared task, this paper describes a very simple method for the annotation and normalization of pharmacological, chemical and, ultimately, biomedical named entities in clinical cases. The system developed for the shared task is based on limited knowledge, collected, structured and munged in a way that clearly outperforms scores obtained by similar dictionary-based systems for English in the past. Along with this recovering of the knowledge-based methods for \NER/ in subdomains,
the paper also highlights the key contribution of \textit{resource-based} systems in the validation and consolidation of both the annotation guidelines and the human annotation practices. In this sense, some of the authors discoverings on the overall quality of human annotated datasets question the above-mentioned `official' results obtained by this system, that ranked second (0.91 F1-score) and first (0.916 F1-score), respectively, in the two \pharmaconer/ subtasks.

\end{abstract}

\section{Introduction}

Named Entity Recognition (\NER/) is considered a necessary first step in the linguistic processing of any new domain, as it facilitates the development of applications showing co-occurrences of domain entities, cause-effect relations among them, and, eventually, it opens the (still to be reached) possibility of understanding full text content. On the other hand, Biomedical literature and, more specifically, clinical texts, show a number of features as regards \NER/ that pose a challenge to NLP researchers~\cite{Cohen_and_Demner-Fushman:2014}: (1) the clinical discourse is characterized by being conceptually very dense; (2) the number of different classes for \NE/s is greater than traditional classes used with, for instance, newswire text; (3) they show a high formal variability for \NE/s (actually, it is rare to find entities in their ``canonical form''); and, (4) this text type contains a great number of ortho-typographic errors, due mainly to time constraints when drafted.

Many ways to approach \NER/ for biomedical literature have been proposed, but they roughly fall into three main categories: rule-based, dictionary-based (sometimes called knowledge-based) and machine-learning based solutions. Traditionally, the first two approaches have been the choice before the availability of Human Annotated Datasets (\HAD/), albeit rule-based approaches require (usually hand-crafted) rules to identify terms in the text, while dictionary-based approaches tend to miss medical terms not mentioned in the system dictionary~\cite{Rebholz-Schumann_et_al:2011}. Nonetheless, with the creation and distribution of \HAD/ as well as the development and success of supervised machine learning methods, a plethora of data-driven approaches have emerged ---from Hidden Markov Models (HMMs)~\cite{Ephraim:2002}, Support Vector Machines (SVMs)~\cite{Habib_and_Kalita:2010} and Conditional Random Fields (CRFs)~\cite{He_and_Kayaalp:2008}, to, more recently, those founded on neural networks~\cite{Armengol-Estapeetal:2019}. This fact has had an impact on knowledge-based methods, demoting them to a second plane. Besides, this situation has been favoured by claims on the uselessness of gazetteers for \NER/ in, for example, Genomic Medicine (GM), as it was suggested by \citeauthor{Cohen_and_Demner-Fushman:2014} \shortcite[p.~26]{Cohen_and_Demner-Fushman:2014}:

\begin{quote}
One of the findings of the first BioCreative shared task was the demonstration of the long-suspected fact that gazetteers are typically of little use in GM.
\end{quote}

 Although one might think that this view could strictly refer to the subdomain of GM and to the past ---BioCreative I was a shared task held back in 2004---, we can still find similar claims today, not only referred to rule-based and dictionary-based methods, but also to stochastic ones~\cite{Armengol-Estapeetal:2019}. \\

In this paper, in spite of previous statements, we present a system that uses rule-based and dictionary-based methods combined (in a way we prefer to call \textit{resource-based}). Our final goals in the paper are two-fold: on the one hand, to describe our system, developed for the \pharmaconer/ shared task\footnote{\url{http://temu.bsc.es/pharmaconer/}}, dealing with the annotation of some of the \NE/s in health records (namely, pharmacological, chemical and biomedical entities) using a revisited version of rule- and dictionary-based approaches; and, on the other hand, to give pause for thought about the quality of datasets (and, thus, the fairness) with which systems of this type are evaluated, and to highlight the key role of resource-based systems in the validation and consolidation of both the annotation guidelines and the human annotation practices. \\

In section~\ref{sect:resources}, we describe our initial resources and explain how they were built, and try to address the issues posed by features (1) and (2) above. Section~\ref{sect:devel} depicts the core of our system and the methods we have devised to deal with text features (3) and (4). Results obtained in \pharmaconer/ by our system are presented in section~\ref{sect:results}. Section~\ref{sect:discussion} details some of our errors, but, most importantly, focusses on the errors and inconsistencies found in the evaluation dataset, given that they may shed doubts on the scores obtained by any system in the competition. Finally, we present some concluding remarks in section~\ref{sect:conclusions}.

\section{Resource building}
\label{sect:resources}

As it is common in resource-based system development, special effort has been devoted to the creation of the set of resources used by the system. These are mainly two ---a flat subset of the \snomed/ medical ontology\footnote{From \url{https://browser.ihtsdotools.org/}.}, and the library and a part of the contextual regexp grammars developed by \citeauthor{FSL:2018} \shortcite{FSL:2018} for a previous competition on abbreviation resolution in clinical texts written in Spanish. The process of creation and/or adaptation of these resources is described in this section.

\subsection{SNOMED CT}
\label{subsect:snomed}

Although the competition proposes two different scenarios, in fact, both are guided by the \snomed/ ontology ---for subtask 1, entities must be identified with offsets and mapped to a predefined set of four classes (\texttt{PROTEINAS}, \texttt{NORMALIZABLES}, \texttt{NO\_NORMALIZABLES} and \texttt{UNCLEAR}); for subtask 2, a list of all \snomed/ \textsc{id}s (\sctid/) for entities occurring in the text must be given, which has been called \textit{concept indexing} by the shared task organizers\footnote{In the train\plus dev datasets, only 17 of the \texttt{PROTEINAS} (`proteins') and \texttt{NORMALIZABLES} (`standardizable') entities have an \textsc{id} not in the \snomed/ ontology. Besides, just 40 out of 5,615 annotations ---not taking into account the class \texttt{UNCLEAR}, which is not considered for the system evaluation--- are tagged as \texttt{NO\_NORMALIZABLES} (`non standardizable'), many of them due to the fact that they include elliptical constructions.}. Moreover, \pharmaconer/ organizers decided to promote \snomed/ substance \textsc{id}s over product, procedure or other possible interpretations also available in this medical ontology for a given entity. This selection must be done even if the context clearly refers to a different concept, according to the annotation guidelines\footnote{\url{https://bit.ly/2qxofgd}, p.~4.}
(henceforth, \AG/) and the praxis. Finally, \texttt{PROTEINAS} is ranked as the first choice for substances in this category.

These previous decisions alone on the part of the organizers greatly simplify the task at hand, making it possible to build (carefully compiled) subsets of the entities to be annotated. This is a great advantage over open domain \NER/, where (like in GM) the texts may contain an infinite (and very creative indeed) number of \NE/s. For clinical cases, although the \NE/ density is greater, there exist highly structured terminological resources for the domain. Moreover, the set of classes to use in the annotation exercise for subtask 1 has been dramatically cut down by the organizers.

With the above-mentioned initial constraints in mind, we have painstakingly collected, from the whole set of \snomed/ terms, instances of entities as classified by the human annotators in the datasets released by the organizers and, when browsing the \snomed/ web version, we have tried to use the ontological hierarchical relations to pull a complete class down from \snomed/. This way, we have gathered 80 classes ---from lipids to proteins to peptides or peptide hormones, from plasminogen activators to dyes to drugs or medicaments---, that have been arranged in a ranked way so as to mimic human annotators choices\footnote{Note that we have gathered the complete set of medical terms included in \snomed/, but, for the purpose of this shared task, we only use a subset of it.}. The number of entities so collected (henceforth, `primary entities') is 51,309.

\subsection{Contextual regexp grammars}
\label{subsect:regexp}

Some of the entities to be annotated, specially those in abbreviated form, are ambiguous without a context. This is the case, for instance, of \textit{PCR}, whose expanded forms are (among other meanings; we use only English expanded forms) `reactive protein c', `polymerase chain reaction', `cardiorespiratory arrest'.
In order to deal with these cases, we use a contextual regexp rule system with a lean and simple rule formalism previously developed~\cite{FSL:2018}. As an exemplification, we include one rule to deal with one of the cases of the preceeding ambiguity:

{\small
\begin{verbatim}
b:[il::bioquímica|en sangre|hemoglobina|
   hemograma|leucocit|parásito|plaqueta|
   prote.na|recuento|urea] - [PCR] - >
[m=proteína]
\end{verbatim}
}

A rule has a left hand side (LHS) and a right hand side (RHS). There is a focus in the LHS (\texttt{PCR}, within dashes) and a left and right context (that may be empty). When the left context includes a \texttt{b:} (like in this case), it indicates either left or right context. The words in the context can take other qualifiers ---in this case, the matching will be case insensitive (\texttt{i} to the left of \texttt{bioquímica}) and local (\texttt{l}), which means the disjunction of words and/or stems can be found in a distance of 40 characters (this can be modifified by the user). Hence, the rule applies, selecting the \texttt{proteína} expansion (in RHS) of \texttt{PCR} if any of the words/stems specified as local context (40 chars maximum) is matched either to the left or right of the focus term (which is usually an abbreviation).

With no tweaking at all for the datasets in \pharmaconer/ competition, the system annotates correctly 18 out of 20 occurrences of \textit{PCR} in the test dataset (a precision of 0.9)\footnote{Note that 2 of the \textit{PCR} occurrences in the train\plus dev datasets have been incorrectly mapped to the protein interpretation (file \texttt{S1130-63432014000100012-1}, 2 times).}.

This component of the system is important because, only when the previous abbreviation is expanded as the first string (that of a protein name), it must be annotated, according to the \AG/. The same ambiguity happens with \textit{Cr}, which may mean `creatinine' or `chrome'\footnote{Again, one of the occurrences of \textit{Cr} has been incorrectly mapped to the former extended form (file \texttt{S0212-16112012000500042-1}).}. These expansions are both \texttt{NORMALIZABLES}, but, obviously, their \sctid/ is different.

The system currently uses 104 context rules, only for abbreviations and acronyms in the clinical cases. These rules, contrary to what is commonly referred in the biomedical processing literature~\cite{Armengol-Estapeetal:2019}, do not require a special domain knowledge (none of the authors do have it) and can be written, most of the times, in a very straightforward way in the formalism briefly described above.

\section{Development}
\label{sect:devel}

In general, dictionary-based methods rely on strict string matching over a fixed set of lexical entries from the domain. This is clearly insufficient to deal with non-canonical linguistic forms of \NE/s as used in clinical texts. For this reason, we have devised two different solutions to this shortcoming.

In the first place, we have munged a great number of our primary entities, in a way similar to that described in \citeauthor{FSL:2019a} \shortcite{FSL:2019a} for gazetteers used for protected information anonymization in clinical texts. We basically transform canonical forms in other possible textual forms observed when working with biomedical texts. With such transformations, a system module converts a salt compound like \textit{clorhidrato de ciclopentolato} into \textit{ciclopentolato clorhidrato}, or simply the \texttt{PP} \textit{de potasio} into its corresponding adjective \textit{potásico}. Other, more complex conversions include the treatment of antibodies ---for instance \textit{anticuerpo contra especie de Leishmania} becomes \textit{ac. Leishmania}, among other variants---, or pairs of antibiotics normally prescribed together ---which have a unique \sctid/ and whose order we handle just as the `glueing' characters. Note, incidentally, that, while the input to this pre-processing step is always a string, the output can be a regular expression, that is linked to a \sctid/. Plural forms are also generated through this module, that uses 45 transformations (not all equally productive). Using these transformation rules, we produce 139,150 `secondary entities', many of them regexps. As a final (simple) example of this, consider the entity \textit{antígeno CD13}: after applying one of the previous string-to-regexp transformations, it is converted to:

{\small
\begin{verbatim}
(?:antígeno )?CD[- ]?13
\end{verbatim}
}

With the previous regexp, the system is able to identify (and string-normalize) six different textual realizations of the same unique \snomed/ term. There are more complex rules that, thus, produce many more potential strings. The important thing with this strategy is that through the generative power of these predictably-created regexps from \snomed/ entities the system is able to improve its recall and overcome the limitations of traditional dictionary-based approaches. \\

Secondly, to tackle with careless drafting of clinical reports, a Levenshtein edit distance library\footnote{We use \texttt{Text::Levensthtein::Flexible} library, from \texttt{Perl} ecosystem.
One of the anonymous reviewers has shed doubts about the use of \texttt{Perl} as a language for ``NLP and text-mining nowadays''. In this respect, we are not committed with a given programming language more than we are with our native language ---and we have submitted our paper in English, a foreign language for us. The system could have been implemented in any other programming language more popular ``nowadays'', provided that we were as \textit{proficient} in it as we are in \texttt{Perl} and the language used were as efficient in string and regexp handling and in I/O operations as \texttt{Perl} is. In this regard, the most popular language nowadays ---\texttt{Python}--- is 2 to 10 times slower for these particular features. \texttt{Perl} is even faster for regexp processing than \texttt{Python PyPy} ---see, for instance, \url{https://github.com/mariomka/regex-benchmark}. Idiomatic \texttt{Perl} is even faster. Finally, \texttt{Perl} has a long tradition in biology and medicine text processing.} is used on the whole background dataset. The process is run once, using our secondary entities as lexicon\footnote{With enumeration of strings from non-infinite-loop regexps.} and a general vocabulary lexicon to rule out common words in the candidate search process. We have used distances in the range 1-3 (depending on string length) for sequences up to 3 words long\footnote{These words are not isolated from the byte stream, and the process uses textual anchors to delimit them as word candidates. Consequently, no proper tokenization is performed.}. The output of this process, which links forms with spelling errors with canonical ones and, thus, to \sctid/s, can be inspected prior to its inclusion in the system lexicon, if so desired. \\

\subsection{Annotation process}

As such, the annotation process is very simple. The program reads the input byte stream trying to identify known entities by means of a huge regexp built through the pre-processing of the available resources. If the candidate entity is ambiguous and (at least) one contextual rule exists for it, it is applied. For the rest of the \NE/s, the system assigns them the class and \sctid/ found in our ranked in-memory lexicon. As already mentioned in passing, the system does not tokenize text prior to \NER/, a processing order that we consider the right choice for highly entity-dense texts. The data structures built during pre-processing are efficiently stored on disk for subsequent runs, so the pre-processing is redone only when resources are edited.

\section{Results}
\label{sect:results}

According to the organizers, and taking into account the \HA/ of the tiny subset from the background dataset released to the participants\footnote{When compared with the rest of the tasks in BioNLP-OST 2019, the time given to \textsf{\scriptsize\textsc{PharmaCoNER}} participants to submit their system runs is 4 times longer than the mean ---longer time that is unnecessary if system is mature enough. On the other hand, the dataset released for evaluation purposes is more than 4 times larger than the mean. As a consequence, participating groups have to annotate full domain corpora rather than just test dataset(s). A shorter submission period and a smaller test dataset would be preferable, and besides fairer, in future calls.}, the system obtained the scores presented in table~\ref{table:results}, ranking as second best system for subtask1 and best system for subtask2~\cite{Agirre:2019}.\footnote{The authors have been unable to obtain these results with the official script, downloaded from \url{https://github.com/PlanTL-SANIDAD/PharmaCoNER-CODALAB-Evaluation-Script}. In their execution of the evaluation script, system results are better (?).}.

\def\pharmaconer/{\textsf{\small\textsc{PharmaCoNER}}}
\begin{table}
\begin{center}
\begin{tabular}{l|r|r|r}
\hline
& Precision & Recall & F1-score \\
\hline
\textbf{Subtask 1} & 0.90625	&	0.91314	&  0.90968 \\
\textbf{Subtask 2} & 0.91108 	&  0.92083	&  0.91593 \\
\hline
\end{tabular}
\caption{Results for \pharmaconer/ test dataset (both subtasks)}
\label{table:results}
\end{center}
\end{table}

Our results are consistent with our poor understanding of the classes for subtask 1. Having a null knowledge of Pharmacology, Biomedicine or even Chemistry, assigning classes (as requested for subtask 1) to entities is very hard, while providing a \sctid/ (subtask 2) seems an easier goal. We will explain the point with an example entity ---\textit{ácido hialurónico} (`hyaluronic acid'). Using the ontological structure of \snomed/, one can find the following parent relations (just in English): \\

\textit{hyaluronic acid} \textsc{\texttt{is-a}} \textit{mucopolysaccharide} \textsc{\texttt{is-a}} \textit{protein} \\

The authors have, in this case, promoted the \texttt{PROTEINAS} annotation for this entity, disregarding its interpretation as a replacement agent and overlooking a recommendation on polysaccharides in the \AG/. Fortunately, all its interpretations share a unique \sctid/. The same may be true for \\

\textit{haemosiderin} \textsc{\texttt{is-a}} \textit{protein} \\

\noindent
which is considered \texttt{NORMALIZABLE} in the test dataset. Similar cases are responsible for the lower performance on subtask 1 with respect to the more complex subtask 2. \\

In spite of these human classification errors, our system scores outperform those obtained by PharmacoNER Tagger\footnote{The tagger authors, some of them also organizers of shared task, have changed the casing of the name for the program.}~\cite{Armengol-Estapeetal:2019}, a simpler system using a binary classification and a very different organization of the dataset with a smaller fragment for test (10\% of the data as opposed to 25\% for the official competition). In fact, our system improves their F1-score (89.06) by 1.3 points when compared with our results for the more complex \pharmaconer/ subtask 1.

\section{Discussion}
\label{sect:discussion}

In this section, we perform error analysis for our system run on the test dataset. We will address both recall and precision errors, but mainly concentrate on the latter type, and on a thorough revision of mismatches between system and human annotations. \\

In general, error analysis is favoured by knowledge-based methods, since it is through the understanding of the underlying reasons for an error that the system could be improved. Moreover, and differently to what happens with the current wave of artificial neural network methods, the whole annotation process ---its guidelines for human annotators, the collection and appropriate structuring of resources, the adequate means to assign tags to certain entities but not to other, similar or even pertaining to the same class--- must be clearly understood by the designer/developer/data architect of such systems. As a natural consequence of this attempt to mimick a task defined by humans to be performed, in the first place, also by humans, some inconsistencies, asystematic or missing assignments can be discovered, and this information is a valuable treasure not only for system developers but also for task organizers, guideline editors and future annotation campaigns, not to mention for the exactness of program evaluation results.

Most of the error types made by the system (i.e., by the authors) in class assignment for subtask 1 have already been discussed. In the same vein, as regards subtask 2, a great number of errors come from the selection of the `product containing substance' reading from \snomed/ rather to the `substance' itself. This is due to inexperience of the authors on the domain and the wrong consideration of context when tagging entities ---the latter being clearly obviated in the \AG/.

In the following paragraphs, some of the most relevant inconsistencies found when performing error analysis of our system are highlighted. The list is necessarily incomplete due to space constraints, and it is geared towards the explanation of our possible errors.

\subsection{Inconsistency in the {\textbf{\sc{ag}}}}
\label{subsect:inconsistency:guide}

Among some of the paradoxical examples in the \AG/ it stands out the double explicit consideration of \textit{gen} (`gene'), when occurs alone in context, as both an entity to be tagged (positive rule P2 of the \AG/) and a noun not to be tagged (negative rule N2). This inconsistency (and a bit of bad luck) has produced that none of the 6 occurrences as an independent noun ---not introducing an entity--- is tagged in the train\plus dev (henceforth, t\plus d) while the only 2 in the same context in the test dataset have been tagged. This amounts for 2 true negatives (\textsc{tn}s) for the evaluation script.

\subsection{Inconsistency in {\textbf{\sc{ha}}} as regards {\textbf{\sc{ag}}}}
\label{subsect:inconsistency:interpreting-guide}

The \AG/ proposal for the treatment of elliptical elements is somewhat confusing. For these cases, a longest match annotation is proposed, which is difficult to replicate automatically and not easy to remember for the human annotator. In many contexts, the annotator has made the right choice ---for instance, in \textit{receptores de estrógeno y de progesterona}--- whereas in others do not ---\textit{$|$anticuerpos anticardiolipina$|$ $|$IgG$|$ e $|$IgM$|$}, with `$|$' marking the edges of the annotations. The last example occurs twice in the test dataset. Hence, the disagreement counts as 6 \textsc{tn}s and 2 false positives (\textsc{fp}s)\footnote{When we indicate this kind of information, mostly using only \textsc{fp}s, it must be understood that the system made the choice(s) that the authors judge as correct, although disagreeing with \HA/ and/or \AG/.}.

On the other hand, there is a clear reference to food materials and nutrition in the \AG/, where they are included in the class of substances. However, none of the following entities is tagged in the test dataset: \textit{azúcar} (which is mandatory according to \AG/ and was tagged in t\plus d; 1 \textsc{fp}); \textit{almidón de maíz} (also mandatory in \AG/; 1 \textsc{fp}); and \textit{Loprofín}, \textit{Aglutella}, \textit{Aproten} (hypoproteic nutrition products, 3 \textsc{fp}s in total)\footnote{On nutrition replacements, see also section~\ref{subsect:inconsistency:train}.}.

There is an explicit indication in the \AG/ to annotate salts, with the example \textit{iron salts}. However, in the context \textit{sales de litio} (`lithium salts'), only the chemical element has been tagged (1 \textsc{fp}\footnote{Note, in passing, that these span errors account for 1 \textsc{tn} also for the evaluation scripts.}).

There exist other differing-span mismatches between human and automatic annotation. These include \textit{anticuerpos anticitoplasma de neutrófilo}, where the \HA/ considers the first two words only (in one of the occurrences, 1 \textsc{fp}); in the text fragment \textit{b2 microglobulina, CEA y CA 19,9 normales}, \textit{CA 19,9} is the correct span for the last entity (and not \textit{CA}, 1 \textsc{fp}); \textit{A.S.T} is the span selected (for \textit{A.S.T.}, 1 \textsc{fp}); finally, in the context \textit{lgM anticore} only \textit{lgM} has been tagged (1 \textsc{fp}).

Other prominent mismatch between \HAD/ and \AG/ is that of \textit{DNA}, which is explicitly included in the \AG/ (sects. P2 and O1). It accounts for 2 \textsc{fp}s. \\

But perhaps one of the most common discrepancies between human and automatic annotation has to do with medicaments normally prescribed together, which have a unique \sctid/. Examples include \textit{amiloride/hidroclorotiazida} (1 \textsc{fp}); and \textit{betametasona + calcipotriol} (1 \textsc{fp}) in the test set. This situation was also observed in the t\plus d corpus fragment (\textit{tenofovir + emtricitabina}, \textit{carbo\-nato cálcico /colecalciferol}, \textit{lopinavir/ritonavir}).

\subsection{Inconsistency in {\textbf{\sc{ha}}} on the test set as regards t\plus d sets}
\label{subsect:inconsistency:train}

Some inconsistencies between dataset annotations have turned the authors crazy: \textit{NPT} (acronym for `total parenteral nutrition, TPN') is tagged in the train\plus dev dataset 15 out of 21 times it occurs\footnote{However, at least one expanded variant of it ---\textit{nutrición parenteral}, `parenteral nutrition'--- is never tagged.}. The common sense of frequency in the \HA/ of texts has led us to tag it in the background set. Unluckily, neither \textit{NPT} nor its expansion have been tagged in the test dataset. This has also been the behaviour in \HA/ for `parenteral nutrition' and `enteral nutrition' (and their corresponding acronyms) in test dataset, since these entities have not been tagged. We asked the organizers about this and other entities for which we had doubts, either because the \AG/ didn't cover their cases or because the \HA/ didn't match the recommendations in the \AG/. Woefully, communication with the organizers has not been very fluent on this respect. All in all, this bad decision on the part of the authors amounts for 6 \textsc{fp}s (more than 7.5\% of our \textsc{fp}s according to evaluation script).

For other cases, decisions that may be clearly induced from the tagging of train\plus dev datasets, have not been applied in the test corpus fragment. These include \textit{cadenas ligeras} (5 times in t\plus d, 1 \textsc{fp} in test); \textit{enzimas hepáticas} (tagged systematically in t\plus d, 1 \textsc{fp}); \textit{p53} (also tagged in t\plus d, 1 \textsc{fp}).

Another entity that stands out is \textit{hidratos de carbono} (`carbohydrates'). It is tagged twice in the t\plus d dataset, occurring 4 times in the set (once as \textit{HC}). However, although the form \textit{carbohidratos} has been annotated twice in the test set, \textit{hidratos de carbono} has been not (1 \textsc{fp}).

Moreover, \textit{suero} (`Sodium chloride solution' or `serum') deserves its own comment. Both entity references are tagged in the train\plus dev datasets (although with the latter meaning it is tagged only 4 out of 12 occurrences). We decided to tag it due to its relevance. In the test dataset, it occurs 5 times with the blood material meaning, but it has only been tagged twice as such (one of them being an error, since it refers to the former meaning). Our system tagged all occurrences, but tagged also one of the instances with the former meaning as serum (3 \textsc{fp}s).

Finally, there are some inconsistencies within the same dataset. For example, nutricional agent \textit{Kabiven} is tagged as both \texttt{NORMALIZABLES} (with \sctid/) and \texttt{NO\_NORMALIZABLES} in the very same text. The same happens with another nutritional complement, \textit{Cernebit}, this time in two different files. The perfusion solution \textit{Isoplasmal G} (with a typo in the datasets ---\textit{Isoplasmar G}) is tagged as \texttt{NORMALIZABLES} and \texttt{UNCLEAR}. These examples reveal a vague understanding (or definition) of criteria as regards fluids and nutrition, as we pointed out at the beginning of this section.

\subsection{Asystematic/incomplete annotation}
\label{subsect:asystematic}

Some of the entities occurring in the test dataset have not always been tagged. This is the case for \textit{celulosa} (annotated only once but used twice, 1 \textsc{fp}); \textit{vimentina} (same situation as previous, 1 \textsc{fp});
\textit{LDH} (tagged 20 times in t\plus d but not in one of the files, 1 \textsc{fp}); \textit{cimetidina} (1 \textsc{fp}); \textit{reactantes de fase aguda} (2 \textsc{fp}s; 2 other occurrences were tagged); \textit{anticuerpos antinucleares} (human annotators missed 1, considered \textsc{fp}).

\subsection{Incorrect {\small\textsf{sctid}s}}
\label{subsect:incorrect}

On our refinement work with the system, some incorrect \sctid/s have emerged. These errors impact on subtask 2 (some also on subtask 1). A large sample of them is enumerated below.

\textit{ARP} (`actividad de renina plasmática', `plasma renin activity', PRA) cannot be linked to \sctid/ for \textit{renina}, which happens twice. In the context `perfil de antigenos [sic] extraíbles del núcleo (ENA)', \textit{ENA} has been tagged with \sctid/ of the antibody (1 \textsc{fp}). In one of the files, \textit{tioflavina} is linked to \sctid/ of \textit{tioflavina T}, but it could be \textit{tioflavina S}. Thus, it should be \texttt{NO\_NORMALIZABLE}. \textit{Harvoni} is {\small \textsf{ChEBI:85082}} and not \textsf{\texttt{<null>}} (1 \textsc{fp}). \textit{AcIgM contra CMV} has a wrong \sctid/ (1 \textsc{fp}). \textit{HBsAg} has no \sctid/ in the test set; it should be {\small\textsf{22290004}} (`Hepatitis B surface antigen') (1 \textsc{fp}). \\

There are other incorrect annotations, due to inadvertent human errors, like \textit{biotina} tagged as \texttt{PROTEINAS} or \textit{VEB} (`Epstein-Barr virus') being annotated when it is not a substance. Among these mismatches between \HA/ and system annotation, the most remarkable is the case of synonyms in active principles. For instance, the brand name drug \textit{Dekapine} has been linked to `ácido valproico' in the former case and to `valproato sódico' in the latter. These terms are synonymous\footnote{Although, `valproato sódico' is the name used in the leaflet, as it can be seen in the Spanish Medicament Agency, AEMPS, web page (\url{https://cima.aemps.es/cima/dochtml/p/
48828/P\_48828.html}: last consulted on 16.07.2019).}, but sadly they don't share \sctid/. Hence, this case also counts as a \textsc{fp}. \\

A gold standard dataset for any task is very hard to develop, so a continuous editing of it is a must\footnote{Besides, when the dataset is being used in a shared task, this refinement process should be available to participants while the task is open.}. In this discussion, we have focused on false positives (\textsc{fp}s) according to the script used for system evaluation, with the main purpose of \textit{understanding} the domain knowledge encoded in the linguistic conventions (lexical/terminological items and constructions) used by health professionals, but also the decisions underlying both the \AG/ and the \HA/ practice.

In this journey to system improvement and authors enlightenment, some inconsistencies, errors, omissions have come up, as it has been reflected in this section, so both the guidelines for and the practice of annotation can also be improved in future use scenarios of the clinical case corpus built and maintained by the shared task organizers.

Our conclusion on this state of affairs is that some of the inconsistencies spotted in this section show that there were not a rational approach to the annotation of certain entities contained in the datasets (apart from other errors and/or oversights), and, hence, the upper bound of any tagging system is far below the ideal 1.0 F1-score. To this respect, in very many cases, the authors have made the wrong choice, but in others they were guided by analogy or common sense. Maybe a selection founded on probability measures estimated on training material could have obtained better results with this specific test dataset. However, in the end, this cannot be considered as an indication of a better system performance, since, as it has been shown, the test dataset used still needs more refinement work to be used as the right dataset for automatic annotation evaluation.

\section{Conclusions}
\label{sect:conclusions}

With this resource-based system developed for the \pharmaconer/ shared task on \NER/ of pharmacological, chemical and biomedical entities, we have demonstrated that, having a very limited knowledge of the domain, and, thus, making wrong choices many times in the creation of resources for the tasks at hand, but being more flexible with the matching mechanisms, a simple-design system can outperform a \NER/ tagger for biomedical entities based on state-of-the-art artificial neural network technology. Thus, knowledge-based methods stand on their own merits in task resolution.

But, perhaps most importantly, the other key point brought to light in this contribution is that a resource-based approach also favours a more critical stance on the dataset(s) used to evaluate system performance. With these methods, system development can go hand in hand with dataset refinement in a virtuous circle that let us think that maybe next time we are planning to add a new gazetteer or word embedding to our system in order to try to improve system performance, we should first look at our data and, like King Midas, turn our Human Annotated Dataset into a true Gold Standard Dataset.

\section*{Acknowledgements}
We thank three anonymous reviewers of our manuscript for their careful reading and their many insightful comments and suggestions. We have made our best in providing a revised version of the manuscript that reflects their suggestions. Any remaining errors are our own responsability.

\bibliography{PharmaCoNER-FSL-AGL_final}
\bibliographystyle{acl_natbib}

\end{document}